\documentclass{article}

    \usepackage[preprint]{neurips_2024}

\usepackage[utf8]{inputenc} 
\usepackage[T1]{fontenc}    
\usepackage{hyperref}       
\usepackage{url}            
\usepackage{booktabs}       
\usepackage{amsfonts}       
\usepackage{nicefrac}       
\usepackage{microtype}      
\usepackage{xcolor}         
\usepackage{multirow}
\usepackage{amsmath}
\usepackage{algorithm}
\usepackage{algpseudocode}
\usepackage[numbers]{natbib}

\title{Residual vector quantization for KV cache compression in large language model}

\author{%
  Ankur Kumar \\
  \texttt{ankurkr@ucla.edu} \\
}

\begin{document}

\maketitle


\begin{abstract}
  KV cache compression methods have mainly relied on scalar quantization techniques to reduce the memory requirements during decoding. In this work, we apply residual vector quantization, which has been widely used for high fidelity audio compression, to compress KV cache in large language models (LLM). We adapt the standard recipe with minimal changes to compress the output of any key or value projection matrix in a pretrained LLM: we scale the vector by its standard deviation, divide channels into groups and then quantize each group with the same residual vector quantizer. We learn the codebook using exponential moving average and there are no other learnable parameters including the input and output projections normally used in a vector quantization set up. We find that a residual depth of 8 recovers most of the performance of the unquantized model. We also find that grouping non-contiguous channels together works better than grouping contiguous channels for compressing key matrix and the method further benefits from a light weight finetuning of LLM together with the quantization. Overall, the proposed technique is competitive with existing quantization methods while being much simpler and results in ~5.5x compression compared to half precision.
  
  \vspace{1mm}
  \centering Code: \url{https://github.com/iankur/vqllm}
\end{abstract}

\section{Introduction}
Efficient decoding in transformer based large language model (LLM) requires caching past key and value vectors, also known as KV cache. The size of KV cache creates a memory bottleneck for storing and loading the cache, specially at long context lengths \cite{fu2024challenges}. Scalar quantization has been effective in compressing KV cache \cite{kivi,hooper2024kvquant,zhao2024atom}. However, in other domains, vector quantization (VQ) is often used for higher compression rate. For example, generative models for images have used vector quantization for efficient training in the latent space \cite{rombach2022high}. In audio domain, residual vector quantization has been used extensively for fast and near-lossless compression of the underlying data \cite{zeghidour2021soundstream, defossez2022high, kumar2024high}. Further, vector quantization is desirable since it decouples numerical precision and compression, which can be useful for training.

Vector quantization has been used in different contexts for LLMs. Retrieval augmented generation uses product vector quantization  to tradeoff accuracy for efficient retrieval. VQ has been used to speed up inference by compressing model weights \cite{baalen2024gptvq, egiazarian2024extreme} or expand model capacity by learning large codebooks \cite{tamkin2023codebook}. A related field is sparse autoencoders \cite{coates2011importance} for LLMs which reconstruct the features from a pretrained model to find interpretable directions \cite{tamkin2023codebook,rajamanoharan2024improving}. However, they require extrememly large codebooks to facilitate interpretability. Recent works have also explored vector quantization for KV cache. \cite{lingle2024transformervq} use vector quantization to compress key and value matrices while training small language models from scratch. \cite{zhang2024kv} apply coupled quantization to compress KV cache using approximate second-order methods to learn the codebooks.

In this work, we explore the application of vector quantization for KV cache compression. In particular, we work with residual vector quantization due to their wide popularity in compressing raw audio. We show that given sufficient residual depth, the technique can be applied to compress KV cache with minor changes to the standard recipe for vector quantization. This is in contrast to existing works on KV cache compression which normally apply heuristics such as quantizing key embeddings across channel dimension \cite{kivi,hooper2024kvquant} or modify the codebook learning procedure \cite{zhang2024kv}.

\section{Method}
\subsection{Vector Quantization}
\cite{vqvae} proposed vector quantization to learn discrete representations for different modalities including images and audio. The idea is to maximize task-specific loss, which is reconstruction in their work, along with vector quantization loss as follows
\begin{align*}
\log p(x | z_q(x)) + ||sg[z_e(x)] - e||_2^2 + \beta ||z_e(x) - sg[e]||_2^2
\end{align*}
where $x$ is input, $z_e(x)$ is the encoder output, $z_q(x)$ is the quantized output, $e$ is embedding in the codebook and $sg$ stands for \emph{stopgradient} operation. The first term in the loss is the reconstruction objective, the middle term is the codebook loss and the last term is the commitment loss. In stead of codebook loss, exponential moving average can also be used to learn the codebook embeddings. The commitment loss encourages the encoder to learn representations which are closer to codebook embedding to avoid the embedding space from growing. Since we work with pretrained LLMs, we set $\beta$ to 0 in all the experiments.

\subsection{Proposed Method}
 For any input embedding $x \in \mathbf{R}^d$, we scale $x$ by its standard deviation, divide the scaled vector into groups of channels, each of size $\widehat{d}$, and quantize each group with the same residual vector quantizer. We discuss two ways to group the channels in section \ref{sec:vq_layer_ablation}. A residual quantizer uses $K$ codebooks, where each codebook $\{C_i\}_{i=1}^K$ has $|C_i|$ codes of size $\widehat{d}$ each. In this work, we do not use any learnable projection matrices for encoding or decoding. For any embedding $z \in \mathbf{R}^{\widehat{d}}$ to be quantized, we obtain the quantized outputs $z_q$ as shown in algorithm \ref{alg:rvq}. We begin with codebook $C_1$ by computing euclidean distance between $z$ and each codebook vector $C_{1j} \in C_1$. We find the nearest code $C_{1\widehat{j}}$ and update $z$ with $z - C_{1\widehat{j}}$. The process is repeated with the remaining codebooks. The output is the sum of all the nearest codes found in the $K$ codebooks multiplied by the standard deviation of the unquantized input $x$. 

\begin{algorithm}
\caption{Residual vector quantization}\label{alg:rvq}
\begin{algorithmic}
\Require $z, \{C_i\}_{i=1}^K$
\State $z_1 \gets z$
\State $z_q \gets 0$
\For{$i \gets 1$ to $K$}
\State $\widehat{j} \gets \arg \min_{j} ||z_i - C_{ij} || $
\State $ z_i^{q} \gets C_{i\widehat{j}} $
\State $ z_q \gets z_q + z_i^{q} $
\State $ z_{i+1} \gets z_i - z_i^{q} $
\EndFor
\State \Return $z_q$
\end{algorithmic}
\end{algorithm}

\section{Experiments} \label{sec:experiments}
Following \cite{hooper2024kvquant}, we apply quantization to the output of key and value projection matrices, which we will refer to as key and value for brevity. Therefore, quantization is applied before RoPE positional encoding \cite{su2024roformer} and each attention block learns 2 residual quantizers, one for key and value each, when quantizing both key and value.
 Unless mentioned, we quantize both key and value, and a single residual quantizer uses $K = 8$ codebooks, where each codebook $C_i$ has $|C_i| = 2048$ codes with each code having $\widehat{d} = 32$ dimensions. All the experiments use 10K random samples from SlimPajama \cite{cerebras2023slimpajama}, which amounts to about 10M tokens, to learn the quantizer codebooks. We initialize each codebook using k-means clustering on the first batch of the corresponding inputs and update them using exponential moving average with a decay factor of 0.99. We use batch size of 64K tokens which is about 150 steps for Llama-3-8b \cite{dubey2024llama} and 180 steps for Mistral-7b \cite{jiang2023mistral} as they use different tokenizers. However, we find that loss does not change much after roughly 50 steps. We evaluate all the models on old-llm-leaderboard \cite{open-llm-leaderboard} tasks: ARC (25-shot) \cite{clark2018think}, HellaSwag (10-shot) \cite{zellers2019hellaswag}, MMLU (5-shot) \cite{hendrycks2020measuring}, TruthfulQA (0-shot) \cite{lin-etal-2022-truthfulqa}, Winogrande (5-shot) \cite{sakaguchi2021winogrande} and GSM8k (5-shot) \cite{cobbe2021training}. We implement the quantization using a simple triton kernel \cite{tillet2019triton} to fuse the $K$ residual steps. We find this to greatly speedup the experiments as it does not materialize the intermediate encodings. All the experiments are run on single A100 GPU (40 GB SXM4) except finetuning and Gemma-7b \cite{team2024gemma} experiments which require higher memory and were run on single H100 GPU (80 GB PCIe). Finetuning experiment uses a constant learning rate of 1e-5 and reduced batch size of 50K tokens. Learning the codebook for the default codebook configuration takes about 1.5 hours for A100 experiments and 2 hours for H100 experiments.

\subsection{Key and value quantization ablation} \label{sec:vq_layer_ablation}
Table \ref{tab:vq_layer_ablation} shows the results for quantizing key and value embeddings with default codebook configuration described above. The first row shows the results for unquantized base Llama-3-8b model. In the second row, we only quantize the key whereas value remains unquantized. Also, contiguous $\widehat{d} = 32$ channels are grouped together. We find that the model has significant drop on MMLU and GSM8k testsets. However, we see improved performance when we group non-contiguous channels which are $\nicefrac{d}{\widehat{d}} = 4$ channels apart (third row). This is different from prior work on coupled quantization \cite{zhang2024kv} which groups the channels contiguosuly and learns different codebook for different group.  We always use contiguous grouping to quantize value embedding (fourth row), which has smaller degradation compared to quantizing key. GSM8k, in particular, has significant gap for key quantization. Finally, when we quantize key and value together, we find a small degradation relative to only key quantization (third row). All these results used frozen LLM weights. However,  we see noticable improvement when we also finetune the weights in all attention blocks along with quantization (last row).
\begin{table}[h]
  \caption{Results for quantizing attention key and value vectors in Llama-3-8b. First row shows results for the unquantized Llama model whereas subsequent rows shows the results for quantizing key or/and value embeddings. We use the default codebook configuration for all the experiments in this table. By default, channels in key embeddings are grouped non-contiguously whereas value channels are grouped contiguously. See \ref{sec:vq_layer_ablation} for details. Finetune means all weights in the attention blocks are finetuned simultaneously with the quantization step.}
  \label{tab:vq_layer_ablation}
  \centering
  \begin{tabular}{ccccccc}
    \toprule
    VQ & ARC & HellaSwag & MMLU & TruthfulQA & WinoGrande & GSM8k\\
    \midrule
    -- & 58.70 & 82.24 & 65.29 & 43.03 & 78.22 & 49.96 \\
    \midrule
    Key (contiguous) & 57.17 & 81.64 & 61.74 & 43.01 & 76.72 & 43.21 \\
    Key & 57.59 & 81.66 & 63.80 & 42.39 & 77.19 & 44.58 \\
    Value & 58.19 & 81.91 & 64.32 & 42.41 & 78.14 & 47.08 \\
    \midrule
    Key and Value & 57.88 & 81.22 & 62.91 & 41.55 & 76.93 & 44.43 \\
    +finetune & 58.36 & 81.58 & 63.77 & 43.51 & 75.69 & 44.35 \\
    \bottomrule
  \end{tabular}
\end{table}

\subsection{Residual codebook size ablation}
We study the effect of different codebook sizes on the model performance using frozen Llama-3-8b. We experiment with number of codebooks $K$ in each quantizer, number of codes $C$ in each codebook and dimension of codebook entry $\widehat{d}$. We experiment with $K \in \{4, 6, 8 \}$, $C \in \{1024, 2048 \}$ and keep $\widehat{d} = 32$ as we saw larger value of $\widehat{d}$ led to drop in performance, which is consistent with findings in the previous work \cite{dai2023emu, esser2024scaling} that smaller code size performs better for code lookup. Therefore, there is a tradeoff between accuracy and compression rate. Results are reported in table \ref{tab:vq_codebook_ablation}. We find that $C = 2048$ performs better than 1024 across all settings. However, the gains diminish for larger values of $K$. At $K=8$, we see small difference between 1024 and 2048 codes for most tasks except GSM8k. We also find $K$ to be the most important factor. A value of $K = 8$ recovers most of the original performance (last row). However, a smaller value such as 6 brings significant drop in performance. The effect is more pronounced on MMLU and GSM8k tasks where we see severe degradation for all cases except $K=8$. Therefore, we use $\hat{d}=32,\ C=2048 \text{ and } K=8$ in all the experiments. We keep the standard deviation, used to scale the input before applying quantization, in half precision which leads to a compression rate of 5.5x for this setting. We note that several existing quantization works normally do not compute performance on GSM8k, which \cite{kivi} refer to as hard generation task requiring several heuristics to maintain performance on the testset. Compared to their method, our technique is much simpler and can be combined with other heuristics in existing works.
\begin{table}[h]
  \caption{Results for different codebook size and depth when quantizing attention key and values with frozen Llama-3-8b model. $\widehat{d}$ is dimension of each codebook entry, C is number of entires in each codebook and K is the number of codebooks in a residual vector quantizer.}
  \label{tab:vq_codebook_ablation}
  \centering
  \begin{tabular}{cccccccccc}
    \toprule
    $\widehat{d}$ & C & K & ARC & HellaSwag & MMLU & TruthfulQA & WinoGrande & GSM8k\\
    \midrule
     -- & -- & -- & 58.70 & 82.24 & 65.29 & 43.03 & 78.22 & 49.96\\
    \midrule
    \multirow{6}{*}{32} & \multirow{3}{*}{1024} & 4 & 46.84 & 71.34 & 43.75 & 38.08 & 62.83 & 00.70 \\
    & & 6 & 54.69 & 78.87 & 56.55 & 40.98 & 71.98 & 32.22\\
    & & 8 & 59.13 & 80.94 & 62.03 & 41.36 & 76.56 & 42.61 \\
     \cmidrule{2-9}
    & \multirow{3}{*}{2048} & 4 & 50.68 & 74.46 & 50.03 & 41.73 & 64.33  & 13.27 \\
    & & 6 & 56.57 & 79.89 & 59.88 & 40.67 & 74.19 & 35.33 \\
    & & 8 & 57.88 & 81.22 & 62.91 & 41.55 & 76.93 & 44.43 \\
    \bottomrule
  \end{tabular}
\end{table}

\subsection{Model ablation}
We ablate the proposed quantization method on different model families to test if the proposed method generalizes. Table \ref{tab:vq_model_ablation} shows the results for base LLama-3-8b, Mistral-7b and Gemma-7b models with the default codebook configuration for quantization. We see a similar trend for performance on all tasks for the three model types. In general, GSM8k has a consistent 4-5\% drop in performance followed by MMLU with around 1.5-2.5\% drop. Moreover, Gemma model sees a higher drop for TruthfulQA when compared to Llama-3 and Mistral models.  
\begin{table}[h]
  \caption{Results for different models when quantizing attention key and values with frozen LLM weights and default codebook configuration (see section \ref{sec:experiments} for more details).}
  \label{tab:vq_model_ablation}
  \centering
  \begin{tabular}{cccccccc}
    \toprule
    Model & VQ & ARC & HellaSwag & MMLU & TruthfulQA & WinoGrande & GSM8k\\
    \midrule
     \multirow{2}{*}{Llama-3-8b} & No & 58.70 & 82.24 & 65.29 & 43.03 & 78.22 & 49.96 \\
      & Yes & 57.88 & 81.22 & 62.91 & 41.55 & 76.93 & 44.43 \\
     \midrule
     \multirow{2}{*}{Mistral-7b} & No & 61.35 & 83.61 & 62.64 & 42.14 & 78.93 & 37.60 \\
      & Yes & 60.41 & 82.96 & 61.15 & 40.64 & 78.3 & 33.74 \\
     \midrule
     \multirow{2}{*}{Gemma-7b} & No & 60.49 & 82.21 & 62.89 & 45.21 & 78.45 & 52.01 \\
      & Yes & 58.19 & 81.61 & 60.55 & 41.69 & 76.16 & 47.61 \\
    \bottomrule
  \end{tabular}
\end{table}

\section{Conclusion}
We showed that residual vector quantization can be applied to compress KV cache in large language models. The proposed method is much simpler and does not use heuristics such as quantizing keys across channel dimension or keep top 1\% outliers in high precision. However, the performance drop on GSM8k even with residual depth of $K = 8$ is a concern. Further, 8 codebooks per residual quantizer may introduce computational efficiency challenges, specially in compute-bound scenarios such as pre-fill phase and large batch decoding. Future works may include ways to reduce the number of codebooks per residual quantizer and analyze the computational efficiency of the proposed method. We hope that some of these challenges can be resolved by integrating the codebook learning during pre-training of large language models.

\bibliographystyle{plain}

\begin{thebibliography}{10}

\bibitem{baalen2024gptvq}
Mart~Van Baalen, Andrey Kuzmin, Markus Nagel, Peter Couperus, Artem Bolshakov, Cedric Bastoul, Eric Mahurin, Tijmen Blankevoort, and Paul Whatmough.
\newblock {GPTVQ}: The blessing of dimensionality for {LLM} quantization.
\newblock In {\em Workshop on Efficient Systems for Foundation Models II @ ICML2024}, 2024.

\bibitem{open-llm-leaderboard}
Edward Beeching, Clémentine Fourrier, Nathan Habib, Sheon Han, Nathan Lambert, Nazneen Rajani, Omar Sanseviero, Lewis Tunstall, and Thomas Wolf.
\newblock Open llm leaderboard.
\newblock \url{https://huggingface.co/spaces/open-llm-leaderboard-old/open_llm_leaderboard}, 2023.

\bibitem{clark2018think}
Peter Clark, Isaac Cowhey, Oren Etzioni, Tushar Khot, Ashish Sabharwal, Carissa Schoenick, and Oyvind Tafjord.
\newblock Think you have solved question answering? {T}ry {ARC}, the {AI2} reasoning {C}hallenge.
\newblock {\em arXiv:1803.05457v1}, 2018.

\bibitem{coates2011importance}
Adam Coates and Andrew~Y Ng.
\newblock The importance of encoding versus training with sparse coding and vector quantization.
\newblock In {\em Proceedings of the 28th international conference on machine learning (ICML-11)}, pages 921--928, 2011.

\bibitem{cobbe2021training}
Karl Cobbe, Vineet Kosaraju, Mohammad Bavarian, Mark Chen, Heewoo Jun, Lukasz Kaiser, Matthias Plappert, Jerry Tworek, Jacob Hilton, Reiichiro Nakano, et~al.
\newblock Training verifiers to solve math word problems, 2021.
\newblock {\em URL https://arxiv. org/abs/2110.14168}, 2021.

\bibitem{dai2023emu}
Xiaoliang Dai, Ji~Hou, Chih-Yao Ma, Sam Tsai, Jialiang Wang, Rui Wang, Peizhao Zhang, Simon Vandenhende, Xiaofang Wang, Abhimanyu Dubey, et~al.
\newblock Emu: Enhancing image generation models using photogenic needles in a haystack.
\newblock {\em arXiv preprint arXiv:2309.15807}, 2023.

\bibitem{defossez2022high}
Alexandre D{\'e}fossez, Jade Copet, Gabriel Synnaeve, and Yossi Adi.
\newblock High fidelity neural audio compression.
\newblock {\em Transactions on Machine Learning Research}, 2023.
\newblock Featured Certification, Reproducibility Certification.

\bibitem{dubey2024llama}
Abhimanyu Dubey, Abhinav Jauhri, Abhinav Pandey, Abhishek Kadian, Ahmad Al-Dahle, Aiesha Letman, Akhil Mathur, Alan Schelten, Amy Yang, Angela Fan, et~al.
\newblock The llama 3 herd of models.
\newblock {\em arXiv preprint arXiv:2407.21783}, 2024.

\bibitem{egiazarian2024extreme}
Vage Egiazarian, Andrei Panferov, Denis Kuznedelev, Elias Frantar, Artem Babenko, and Dan Alistarh.
\newblock Extreme compression of large language models via additive quantization.
\newblock In {\em Forty-first International Conference on Machine Learning}, 2024.

\bibitem{esser2024scaling}
Patrick Esser, Sumith Kulal, Andreas Blattmann, Rahim Entezari, Jonas M{\"u}ller, Harry Saini, Yam Levi, Dominik Lorenz, Axel Sauer, Frederic Boesel, et~al.
\newblock Scaling rectified flow transformers for high-resolution image synthesis.
\newblock In {\em Forty-first International Conference on Machine Learning}, 2024.

\bibitem{fu2024challenges}
Yao Fu.
\newblock Challenges in deploying long-context transformers: A theoretical peak performance analysis.
\newblock {\em arXiv preprint arXiv:2405.08944}, 2024.

\bibitem{hendrycks2020measuring}
Dan Hendrycks, Collin Burns, Steven Basart, Andy Zou, Mantas Mazeika, Dawn Song, and Jacob Steinhardt.
\newblock Measuring massive multitask language understanding.
\newblock {\em Proceedings of the International Conference on Learning Representations (ICLR)}, 2021.

\bibitem{hooper2024kvquant}
Coleman Hooper, Sehoon Kim, Hiva Mohammadzadeh, Michael~W Mahoney, Yakun~Sophia Shao, Kurt Keutzer, and Amir Gholami.
\newblock K{VQ}uant: Towards 10 million context length {LLM} inference with {KV} cache quantization.
\newblock {\em arXiv preprint arXiv:2401.18079}, 2024.

\bibitem{jiang2023mistral}
Albert~Q Jiang, Alexandre Sablayrolles, Arthur Mensch, Chris Bamford, Devendra~Singh Chaplot, Diego de~las Casas, Florian Bressand, Gianna Lengyel, Guillaume Lample, Lucile Saulnier, et~al.
\newblock Mistral 7b.
\newblock {\em arXiv preprint arXiv:2310.06825}, 2023.

\bibitem{kumar2024high}
Rithesh Kumar, Prem Seetharaman, Alejandro Luebs, Ishaan Kumar, and Kundan Kumar.
\newblock High-fidelity audio compression with improved rvqgan.
\newblock {\em Advances in Neural Information Processing Systems}, 36, 2024.

\bibitem{lin-etal-2022-truthfulqa}
Stephanie Lin, Jacob Hilton, and Owain Evans.
\newblock {T}ruthful{QA}: Measuring how models mimic human falsehoods.
\newblock In {\em Proceedings of the 60th Annual Meeting of the Association for Computational Linguistics (Volume 1: Long Papers)}, pages 3214--3252, Dublin, Ireland, May 2022. Association for Computational Linguistics.

\bibitem{lingle2024transformervq}
Lucas~Dax Lingle.
\newblock Transformer-{VQ}: Linear-time transformers via vector quantization.
\newblock In {\em The Twelfth International Conference on Learning Representations}, 2024.

\bibitem{kivi}
Zirui Liu, Jiayi Yuan, Hongye Jin, Shaochen Zhong, Zhaozhuo Xu, Vladimir Braverman, Beidi Chen, and Xia Hu.
\newblock {KIVI}: A tuning-free asymmetric 2bit quantization for {KV} cache.
\newblock In {\em Proceedings of the 41st International Conference on Machine Learning}, volume 235 of {\em Proceedings of Machine Learning Research}, pages 32332--32344. PMLR, 21--27 Jul 2024.

\bibitem{rajamanoharan2024improving}
Senthooran Rajamanoharan, Arthur Conmy, Lewis Smith, Tom Lieberum, Vikrant Varma, J{\'a}nos Kram{\'a}r, Rohin Shah, and Neel Nanda.
\newblock Improving dictionary learning with gated sparse autoencoders.
\newblock {\em arXiv preprint arXiv:2404.16014}, 2024.

\bibitem{rombach2022high}
Robin Rombach, Andreas Blattmann, Dominik Lorenz, Patrick Esser, and Bj{\"o}rn Ommer.
\newblock High-resolution image synthesis with latent diffusion models.
\newblock In {\em Proceedings of the IEEE/CVF conference on computer vision and pattern recognition}, pages 10684--10695, 2022.

\bibitem{sakaguchi2021winogrande}
Keisuke Sakaguchi, Ronan~Le Bras, Chandra Bhagavatula, and Yejin Choi.
\newblock Winogrande: An adversarial winograd schema challenge at scale.
\newblock {\em Communications of the ACM}, 64(9):99--106, 2021.

\bibitem{cerebras2023slimpajama}
Daria Soboleva, Faisal Al-Khateeb, Robert Myers, Jacob~R Steeves, Joel Hestness, and Nolan Dey.
\newblock {SlimPajama: A 627B token cleaned and deduplicated version of RedPajama}.
\newblock \url{https://cerebras.ai/blog/slimpajama-a-627b-token-cleaned-and-deduplicated-version-of-redpajama}, 2023.

\bibitem{su2024roformer}
Jianlin Su, Murtadha Ahmed, Yu~Lu, Shengfeng Pan, Wen Bo, and Yunfeng Liu.
\newblock Roformer: Enhanced transformer with rotary position embedding.
\newblock {\em Neurocomputing}, 568:127063, 2024.

\bibitem{tamkin2023codebook}
Alex Tamkin, Mohammad Taufeeque, and Noah~D Goodman.
\newblock Codebook features: Sparse and discrete interpretability for neural networks.
\newblock {\em arXiv preprint arXiv:2310.17230}, 2023.

\bibitem{team2024gemma}
Gemma Team, Thomas Mesnard, Cassidy Hardin, Robert Dadashi, Surya Bhupatiraju, Shreya Pathak, Laurent Sifre, Morgane Rivi{\`e}re, Mihir~Sanjay Kale, Juliette Love, et~al.
\newblock Gemma: Open models based on gemini research and technology.
\newblock {\em arXiv preprint arXiv:2403.08295}, 2024.

\bibitem{tillet2019triton}
Philippe Tillet, Hsiang-Tsung Kung, and David Cox.
\newblock Triton: an intermediate language and compiler for tiled neural network computations.
\newblock In {\em Proceedings of the 3rd ACM SIGPLAN International Workshop on Machine Learning and Programming Languages}, pages 10--19, 2019.

\bibitem{vqvae}
Aaron Van Den~Oord, Oriol Vinyals, et~al.
\newblock Neural discrete representation learning.
\newblock {\em Advances in neural information processing systems}, 30, 2017.

\bibitem{zeghidour2021soundstream}
Neil Zeghidour, Alejandro Luebs, Ahmed Omran, Jan Skoglund, and Marco Tagliasacchi.
\newblock Soundstream: An end-to-end neural audio codec.
\newblock {\em IEEE/ACM Transactions on Audio, Speech, and Language Processing}, 30:495--507, 2021.

\bibitem{zellers2019hellaswag}
Rowan Zellers, Ari Holtzman, Yonatan Bisk, Ali Farhadi, and Yejin Choi.
\newblock Hellaswag: Can a machine really finish your sentence?
\newblock In {\em Proceedings of the 57th Annual Meeting of the Association for Computational Linguistics}, 2019.

\bibitem{zhang2024kv}
Tianyi Zhang, Jonah Yi, Zhaozhuo Xu, and Anshumali Shrivastava.
\newblock Kv cache is 1 bit per channel: Efficient large language model inference with coupled quantization.
\newblock {\em arXiv preprint arXiv:2405.03917}, 2024.

\bibitem{zhao2024atom}
Yilong Zhao, Chien-Yu Lin, Kan Zhu, Zihao Ye, Lequn Chen, Size Zheng, Luis Ceze, Arvind Krishnamurthy, Tianqi Chen, and Baris Kasikci.
\newblock Atom: Low-bit quantization for efficient and accurate llm serving.
\newblock {\em Proceedings of Machine Learning and Systems}, 6:196--209, 2024.

\end{thebibliography}

\end{document}